# Application of Segment Anything Model for Civil Infrastructure Defect Assessment

Mohsen Ahmadi[1], Ahmad Gholizadeh Lonbar[2,3], Hajar Kazemi Naeini[4], Ali Tarlani Beris[5,6], Mohammadsadegh Nouri[7,3], Amir Sharifzadeh Javidi[3], Abbas Sharifi[8,*]

*1- Department of Electrical Engineering and Computer Science, Florida Atlantic University, FL, USA*
*2- Department of Civil, Architectural and Environmental Engineering, Missouri University of Science and Technology, Rolla, MO, USA*
*3- School of Civil Engineering, College of Engineering, University of Tehran, Iran*
*4- Department of Civil Engineering, University of Texas at Arlington, Arlington, Texas, USA*
*5- Department of Mechanical Engineering, College of Engineering, Boston University, Boston, MA, USA*
*6- School of Mechanical Engineering, College of Engineering, University of Tehran, Tehran, Iran*
*7- Center for Applied Coastal Research, Department of Civil and Environmental Engineering, University of Delaware, Newark, DE, USA*
*8- Department of Civil and Environmental Engineering, Florida International University, Miami, FL, USA*
*\*Corresponding author: ashar092@fiu.edu*

## Abstract

This research assesses the performance of two deep learning models, SAM and U-Net, for detecting cracks in concrete structures. The results indicate that each model has its own strengths and limitations for detecting different types of cracks. Using the SAM's unique crack detection approach, the image is divided into various parts that identify the location of the crack, making it more effective at detecting longitudinal cracks. On the other hand, the U-Net model can identify positive label pixels to accurately detect the size and location of spalling cracks. By combining both models, more accurate and comprehensive crack detection results can be achieved. The importance of using advanced technologies for crack detection in ensuring the safety and longevity of concrete structures cannot be overstated. This research can have significant implications for civil engineering, as the SAM and U-Net model can be used for a variety of concrete structures, including bridges, buildings, and roads, improving the accuracy and efficiency of crack detection and saving time and resources in maintenance and repair. In conclusion, the SAM and U-Net model presented in this study offer promising solutions for detecting cracks in concrete structures and leveraging the strengths of both models that can lead to more accurate and comprehensive results.

## 1. Introduction

There are various types of damage that civil infrastructure can experience during and after its construction, such as deterioration processes, aging, and diverse loads. Throughout the 21st century, civil organizations have grappled with the challenge of actively managing civil infrastructure in order to prevent unforeseen structural collapses and maintain robust infrastructure. This challenge is compounded by an increasing number of aging buildings, limited material and environmental resources, and constrained financial budgets. According to the American Society of Civil Engineers (ASCE), 39% of American bridges are older than 50 years, and the anticipated cost of maintaining them is $123 billion [1]. The construction maintenance markets in the United Kingdom and Switzerland accounted for approximately 50% of the total value of the construction markets[1]. It is estimated that approximately one-third of the bridge infrastructure throughout Europe was built after World War II and is now out of service. Around 270,000 structures across the Asian require regular inspection each year, but both the funds and the number of inspectors are dwindling [2]. Infrastructure inspections must be conducted continuously and periodically to prevent hazardous working conditions and dangers, necessitating the development of accurate and effective inspection techniques.. Traditional inspection methods involve hiring a qualified inspector with extensive experience to conduct on-site investigations using nondestructive tests, images, annotations, drawings, and historical documentation. However, such inspections are often costly, complex, hazardous, and time-consuming, particularly for structures that are difficult to access. Furthermore, the subjectivity of inspectors and human error may result in varying degrees of inaccuracies over time, depending on the inspector's abilities. Automated health monitoring of civil infrastructures could replace human inspections, addressing the aforementioned concerns while making inspections more cost-effective and achieving higher spatial resolution.

In recent years, vision-based methods for structural health assessment have gained popularity as visual sensing technology (such as digital cameras) and computational models have advanced. There are numerous benefits associated with these state-of-the-art methods, including low cost, high resolution, reliability, and ease of use [2]. Numerous structural issues, including fractures, flaking, rust, and separation, can be identified by examining images of structural exteriors. The benefits of utilizing computer vision techniques encompass distant, touchless, cost-effective, unbiased, and automated assessments of the state [3,4]. Cha et al. [5] successfully identified structural crack damage from images using convolutional neural networks. In another example, Yeum et al. developed a method for filtering out a small fraction of relevant images based on an aerial vision sensor platform [6]. There are various types of computer vision techniques, ranging from traditional methods that rely on image processing algorithms to more recent trends that utilize deep learning algorithms. Traditional detection methods require manual derivation of features that transform available data into useful information. These methods include statistically based approaches for greyscale distributions [7], color and texture descriptors [8], binarization techniques [9], and machine learning-based models [10]. Nonetheless, these algorithms overlook the contextual data offered by areas around the flaws, which restricts the use of image processing in automated structural evaluation settings. These strategies need manual adjustment depending on the target structures requiring inspection. Moreover, oblique long-range images or changes in lighting and shadow parameters during image capture can yield inaccurate results. Due to the diversity of real-world conditions, developing a general algorithm that performs effectively in these scenarios can be challenging. Spencer et al. [11] showcase the application of recent developments in computer vision methods to assess the condition of civil infrastructure. They explore pertinent studies in computer vision, machine learning, and structural engineering fields. In the examined work, the authors categorized inspection and monitoring applications, and pinpointed the crucial obstacles that must be tackled for successful automated vision-driven inspection and monitoring of civil infrastructure. Nowadays, pre-trained models play a significant role in science, technology, and engineering. The emergence of large language models pre-trained with web-scale datasets has revolutionized segmentation [12]. These models, also known as foundation models, can perform generalized tasks beyond those they have been trained for [13]. Empirical studies and observations demonstrate a clear improvement in the performance of these models as the model scale increases and the dataset size and total training computer grow [14]. This trend also applies to computer vision and language encoding, which are among the primary applications of foundation models. Despite the remarkable progress in computer vision and language encoding development, there are still many technical gaps in these fields, primarily due to the lack of abundant training data. Recently, Meta AI Research introduced a novel image segmentation network with an inspiring name, Segment Anything (SA) [15]. This comprehensive project presents a new task, model, and dataset for image segmentation, utilizing the largest segmentation dataset in computer vision to date. The Segment Anything Model (SAM) has been trained with over one billion masks on 11 million images [15]. This massive amount of data can potentially push SAM significantly forward in terms of performance and accuracy compared to existing models. SAM has been developed by addressing three main components: task, model, and data. These three components are intertwined, and a holistic solution is needed to address them. First, a segmentation task is defined is general enough to offer a robust pre-training objective and cover a wide range of applications. This task necessitates a model that supports flexible prompting. Next, the model is trained with a diverse and large-scale dataset. Due to the unavailability of such data sources, a data engine has been created, and an iterative process has been implemented to train the model. This iterative process employs the model itself to automatically collect new data, which is then used to further train and strengthen the model [16-18]. Without a doubt, the dataset used to train SAM is its most distinctive feature.

The final dataset is named SA-1B, indicating 1 billion masks on 11 million images used to train the model. The diversity and high quality of this exceptional dataset ensure a fair and bias-free model. Additionally, an experimental phase has been conducted to test and evaluate SAM's performance, revealing strong quantitative and qualitative results on various tasks under a zero-shot detection protocol, including edge detection, object proposal generation, instance segmentation, and an initial exploration of text-to-mask prediction [19]. SAM has been released as a permissive open-license package in Python, available for public use. In this study, we aim to use a pre-trained Segment Anything Model (SAM) for detecting damage in civil infrastructure. To assess the performance of the SAM, we train a U-Net Convolutional Neural Network (CNN) and compare its output to that of several pre-trained models, including Resnet-18, Resnet-50, MobileNetV2, and Xception. The comparison is done quantitatively, meaning that we compare the numerical results of the different models to determine which one performs the best. The objective of this comparison is to evaluate the performance of our presented method alongside SAM for damage detection in civil infrastructure and to ascertain whether it surpasses other models.

## 2. Related Work

In recent years, deep learning algorithms have improved vision-based damage detection [20]. In this way, a knowledge-driven approach is replaced with a data-driven approach, and human errors are replaced with system errors by automatically extracting features. Damage detection has been applied to a variety of structures and types of faults, including cracks, spalling, and corrosion. In order to offer a classifier for spotting cracks in concrete surfaces, road pavement, and steel box girders, convolutional neural network (CNN) architectures have been deployed [21–23]. Quqa et al. [24] propose a two-step method for determining cracked pixels that identifies cracked areas and then uses image processing techniques to determine broken pixels based on the cracked areas. Yessoufou and Zhu [25] outlined the application of the proposed damage detection method (One-Class Convolutional Neural Network) to monitoring the condition of bridges in practice. Spalling and cracking of concrete surfaces have been classified using the popular AlexNet architecture [26, 27]. Cha et al. [28] and Cha and Choi [29] proposed a vision-based method for detecting concrete cracks using a $256 \times 256 \times 3$ CNN classifier and sliding window technique. They trained the model with a large-scale dataset of 34,680 images using the VGG16 network, achieving an accuracy of 98.8% on the test set and promising results in detecting cracks in complex and diverse environments.

A study conducted by Sandra et al. [30] revealed that the VGG16 model offers the best performance for the detection of multiclass concrete damage, with an average IoU value of 59.5% for delaminations and 39.4% for cracks under thermographic imaging. According to Cha et al. [28] and Cha and Choi [29], the vision-based method used for identifying concrete fractures involves a 256 x 256 x 3 CNN classifier and sliding window methodology. A large dataset of 34,680 images was used to train the model, which showed promising results for the detection of fractures in a diverse and intricate environment. Sandra et al. [30] employed thermographic images to find that the VGG16 model performed the best at identifying multiclass concrete defects, with an average IoU value of 59.5% for delaminations and 39.4% for cracks. Savino and Tondolo [31] compared the classification accuracy of the GoogLeNet architecture to eight different pre-trained networks and obtained a 94% accuracy. Kruachottikul et al. [32] developed an inspection method that detects cracking, erosion, honeycomb, scaling, and spalling in reinforced concrete bridge substructures with 81% accuracy. Since image classification approaches are limited to distinguishing between images based on the intended class, object detection techniques have recently been employed to identify and locate various damages within bounding boxes. According to Cha et al. [27], they identified five different types of concrete damage, including steel corrosion, bolt corrosion, and steel delamination, using a faster Region-based Convolutional Neural Network (R-CNN). Through transfer learning, 2,366 images (each 500 by 375 pixels) of concrete structures with various defects were analyzed to train the R-CNN model. A probability of 87.8% of concrete damage can be determined by the method. It is not possible to completely define the geometry of damages using object detection algorithms, despite faster R-CNNs' success in locating damage in crane projects, urban shield tunnel linings, and historic masonry structures. According to Zhu and Song [33], a weakly supervised segmentation network was created using an autoencoder and k-means clustering for detecting different types of features associated with fractures within an asphalt concrete bridge deck. Due to the dark color, complexity, and variety of problems associated with asphalt concrete bridge decks, conventional crack detection methods cannot accurately and efficiently identify these faults. The study concluded that segmentation is a more effective method for detecting faults than current approaches.

Another study by Li et al. [34] illustrated that a modified Fully Convolutional Network (FCN) could detect cracks, spalling, exposed rebar, and holes in concrete structures through semantic segmentation. They combined the model with a sliding window technique to handle larger images. Their model outperformed traditional edge detection methods in complex environments without needing manual feature extraction. Dung et al. [35] employed a deep Fully Convolutional Network (FCN) for the semantic segmentation of images of concrete cracks. They estimated that they achieved an accuracy of approximately 90%. The proposed approach was validated using a cyclic loading test on a concrete specimen. The results indicated that the model could detect cracks and estimate their density accurately. Kim et al. [36] developed a model to detect concrete cracks using a hierarchical CNN and a multi-loss update segmentation technique, which can be used for structural safety inspections. Shi et al. [37] increased segmentation accuracy using two techniques of data input, squash segmentation and crop segmentation. They found that at a Background Data Drop Rate (BDDR) of 0.8, cropping segmentation becomes comparatively more accurate in terms of accuracy of damaged pixels in the data. Zhang et al. [38] used a context-aware deep semantic segmentation network to identify concrete cracks. According to the results, BF scores improved substantially when compared to state-of-the-art techniques, with Boundary F1 (BF) scores of 0.8234, 0.8252, and 0.7937 for CFD, TRIMMD, and CFTD, respectively. In order to identify flaws in civil infrastructure, Hsu et al. [39] used Deeplabv3+ and pre-trained neural network weights. As part of their study, they assessed the effectiveness of Mask R-CNN and several deep neural network models for detecting and validating cracks.

## 3. Method and Materials

There are different pre-trained segmentation models for damage detection in civil infrastructures. These models are built based on multi architectures, including Resnet-18 and Resnet-50, MobileNetV2, Xception, U-Net, and Segment anything that had already been trained. These pre-trained models can be fine-tuned to specific datasets to improve their accuracy for a particular task. In this research, we aim to evaluate the performance of a pre-trained Segment Anything Model (SAM) in detecting damage to civil infrastructure. To do this, we train a U-Net Convolutional Neural Network (CNN) and compare its results to those of several other pre-trained models, including ResNet-18, ResNet-50, MobileNetV2, and Xception. Our final step has been to compare the results of the testing of the SAM with our proposed network for detecting damage to civil infrastructure. However, we introduce the networks used in this study in this section.

## 3.1. Dataset

To train neural networks, images were previously captured in nearly perfect laboratory settings, with proper camera angles and locations modified according to the appearance and location of faults [26]. The datasets' hundreds of images were also cropped to produce considerably smaller versions of the original images. Since it is difficult to reproduce perfect conditions and continuously adjust the location, angle, and illumination direction of cameras mounted on UAVs, this technique would not be capable of addressing a wide variety of flaws in the field. Since most image-based techniques are based on the data from one experiment, they cannot be applied to other datasets. Numerous research projects have been conducted on the semantic segmentation of single flaws. It has never been investigated whether pre-trained semantic segmentation algorithms can identify multiclass concrete flaws.

Accordingly, the purpose of this study is to propose a CNN that, based on these gaps, can perform semantic segmentation of images comprised of cracks, delamination, and backgrounds in various civil infrastructures. In the first step, a neural network was trained that was able to perform a variety of on-field inspections regardless of the quality of the images. According to Savino et al. [26], 1,250 images were gathered to build the neural network. This image set illustrates genuine environmental circumstances with a variety of backdrops and various sources of noise. A second objective was to find a pre-trained neural network most suitable for spotting civil engineering flaws. In addition to corrosion, efflorescence, stain wetness, and vacancies, additional forms of structural degradation may be identified through further investigation. A morphological analysis was conducted as well to demonstrate how semantic segmentation outperforms all other approaches to finding flaws in civil infrastructures.

## 3.2. Database construction

A neural network's performance is influenced by the variety of training images. The performance of a CNN could be adversely affected if the training dataset only contains images with specific circumstances. To ensure the training dataset reflects a wide range of potential conditions, several raw images were gathered from a variety of sources, including the Internet, on-site inspections of bridges, and Google Street View. As a result of the images collected from three different sources with variable quality, resolutions, and backgrounds, their study is more successful when used with inexpensive sensors in on-field applications. A total of 1,250 images were manually labeled with the labels "Delamination," "Crack," and "Background" using the MATLAB Image Labeler program. Examples of the raw images used to create the dataset are shown in Figure 1, along with the ground truth annotations.

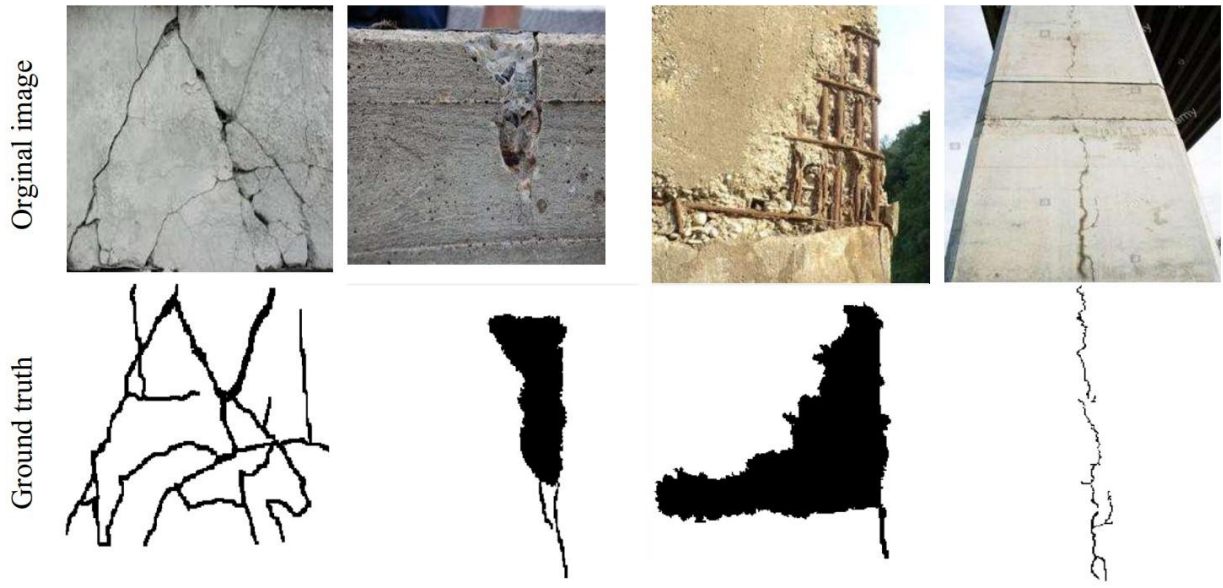

**Figure 1:** Samples of the Original and Ground Truth images

It is important to note that a neural network's performance is influenced by the diversity of its training images. There is a possibility that a CNN's performance may be adversely affected if the training dataset consists only of images that meet certain conditions. Various raw images were acquired from different sources to ensure that the training dataset includes a wide range of possible situations, including images obtained from the Internet, on-site inspections of bridges, and Google Street View. As a result of tagging the images and cropping them to a 300x300 pixel size, computing costs and training time were reduced.

Randomized data augmentation was applied, which included rotations, reflections, and shears, to increase the number of training data in order to enhance the neural networks' resilience to distortions in the image data and reduce the likelihood of overfitting. After data augmentation, the new database was divided into two parts: 80% for the training set (2,000 images) and 20% for the validation set (500 images). In concrete damage datasets, an imbalance in the distribution of class pixels between cracks and other locations is a common issue since fractures often occupy a smaller area. As a result of an imbalanced dataset, the overrepresented classes' errors contribute significantly more than the underrepresented classes' errors, which negatively impacts the performance of the underrepresented classes. During the training phase, class weighting was applied to modify the significance of each pixel in order to resolve this problem and prevent semantic segmentation bias towards dominant classes. The median frequency weighting technique calculates the weight of each class by dividing the total number of pixels in all the images within that class by the number of pixels in each individual image in that class.

**Table 1 :** Pixels and median weights for each image class [18]

|  | **Delamination** | **Crack** | **Background** |
|---|---|---|---|
| Pixel count | $1.4801 \times 10^7$ | $4.3806 \times 10^6$ | $1.3743 \times 10^8$ |
| Frequency | 0.1620 | 0.0273 | 0.7635 |
| Class weight | 1 | 5.9286 | 0.2122 |

Table 1 summarizes the distribution of pixels within each class in the training set. It is the proportion of pixels in an image belonging to each class that determines the frequency of that class. Class weights are assigned according to the frequency of each class and reflect the importance of each class in the training process. Class weighting is used to balance the influence of each class during the training process and to ensure that the neural network does not become biased in favor of the dominant class.

### 3.3. Cracks and their types in concrete structures

The most noticeable defect in concrete structures is cracking. Therefore, the inspection process for concrete structures begins with imaging surfaces as a preliminary step for detecting different types of cracks. Although all types of cracks may occur anywhere in the structure and reduce its durability and load capacity, they can be classified into two major categories based on their visual appearance: (1) longitudinal cracks and (2) spalling cracks. Longitudinal cracks are sharp-edged cracks running parallel to the length of the concrete element. These cracks can also be subcategorized based on their cause of formation.

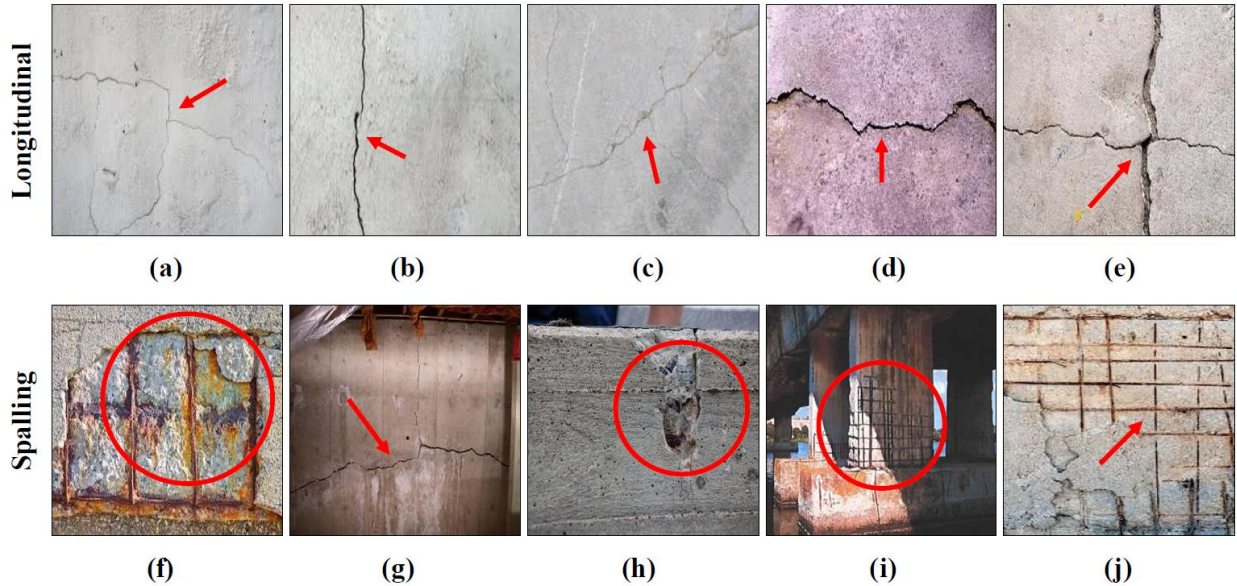

**Figure 2:** Examples of longitudinal and spalling cracks in concrete structures: (a) drying shrinkage cracks, (b) thermal cracks, (c) plastic shrinkage cracks, (d) settlement cracks, (e) overloading cracks, (f) corrosion-induced cracks, (g) alkali-silica reaction cracks, (h) freeze-thaw cracks, (i) chemical attack cracks, and (j) abrasion cracks.

Drying shrinkage cracks, thermal cracks, plastic shrinkage cracks, settlement cracks, and overloading cracks are all longitudinal cracks forming due to the loss of moisture from the concrete, temperature changes in the concrete, rapid drying of the concrete surface, differential settlement of the foundation, and excessive loading on the structure, respectively. On the other hand, spalling cracks are surface defects occurring when the top layer of concrete begins to break away or flake off. Just like longitudinal cracks, spalling cracks are divided into subcategories of Corrosion-induced cracks, alkali-silica reaction cracks, freeze-thaw cracks, chemical attack cracks, and abrasion cracks, which form due to corrosion of the reinforcement (predominantly steel), reactions between the alkalis in the cement and reactive silicas in the aggregates, expansion of water trapped in cavities when it freezes, chemical attacks on the concrete, and abrasion of the concrete surface, respectively. While both types of cracks can be an alarm sign of underlying problems with the concrete structure, spalling cracks are generally superficial and primarily affect the appearance of the concrete, whereas longitudinal cracks can indicate more serious structural issues and may require closer attention and repair. Figure 2 shows examples of different types of cracks in concrete structures.

### 3.4. CNN architecture

CNNs are primarily based on the assumption that the input will be composed of images. The architecture is therefore designed in a manner that makes it particularly suitable for handling a specific kind of data based on this features. There are several key differences between a CNN and a traditional network, including the three-dimensional organization of neurons within the layers, the spatial dimensionality of the input (height and width), and the depth of the input. ANNs do not have a third dimension, depth, which represents the number of layers.

Neurons in each layer of the artificial neural network will only be connected to a small portion of the layer preceding it, as opposed to conventional artificial neural networks. Essentially, this corresponds to an input volume of 64 x 64 x 3 (height, width, and depth), followed by a final output layer of 1 x 1 x n (where n is the number of classes that may be created). Due to the reduction in the input dimensionality, a more manageable number of class scores has been dispersed over the depth dimension.

### 3.5. CNN Segmentation Methods

Over the last few decades, a number of models have been introduced for image segmentation, including thresholding, edge detection, clustering, region-growing, and more sophisticated active contour models for CNN segmentation. In the early days, thresholding and region-growing were possible, but their performance was limited because they only used intensity or texture information from images. [40]. Based on Chan and Vese's work [41], level set functions are introduced as a method for analyzing segmentation problems as partial differential equations (PDEs). In the following years, this model was extended to multiphase problems as well as texture problems [41, 42]. Computational efficiency can be improved by using efficient solvers, such as dual projections and graph cut methods. Models of this type commonly face a number of challenges, including lengthy development periods. Additionally, neural network-based or support vector machine-based supervised segmentation models produced reasonable results. However, as these models rely on handcrafted segmentation features, their applicability and quality are limited. CNNs, a class of deep neural networks, are well-suited for performing various computer vision tasks, such as classification, segmentation, and registration. CNN-based models offer end-to-end functionality and are capable of extracting hierarchical and multi-resolution features. Numerous developments in CNN architectures have occurred, such as AlexNet [42], VGGNet [43], GoogLeNet [44], and DenseNet [45].

CNN-based segmentation models can be categorized into two broad categories: pixel-based and image-based. The pixel-based approach treats each pixel as a classification problem and classifies it into different objects. Patches are often generated for each pixel (or super-pixel) as input to CNN models for classification, with the pixel label serving as the target in the training process [46]. Image-based approaches, such as U-Net [47], use images as inputs and produce segments of the input images (of the same size) as outputs. U-Net-like models have gained popularity due to their superior performance and simplicity compared to pixel-wise approaches [47, 48, 49]. However, small segmented objects may occur around boundaries due to the lack of consideration for features outside the target. To address this issue, Jégou et al. [50] developed a network based on DenseNet called Tiramisu, which contains one hundred layers. Each layer is connected in a feedforward manner, enabling DenseNet to reduce the influence of external features on targets by encouraging feature reuse and strengthening feature propagation.

### 3.6. Overall architecture

Deep Learning is the process of learning more complex concepts from simpler ones using multi-layer artificial neural networks. In deep learning models, learning occurs through the manipulation of uncertainty about predictions of unavailable data using experience-based data. It is like how a human inspector acquires expertise during occupational training in order to assess damage to infrastructure. Convolutional neural networks (CNNs) are among the most commonly used deep learning models, and they are highly effective in image classification tasks. These models have been applied in a broad range of real-world scenarios, from self-driving vehicles to face detection in surveillance systems. CNN models typically consist of layers of convolution and pooling. To implement convolutional layers, input images are multiplied by small feature matrices, which are derived from the input images (corners, edges, sharp intensity changes, etc.), and the sum is normalized based on the size of the matrices (i.e., kernel size).

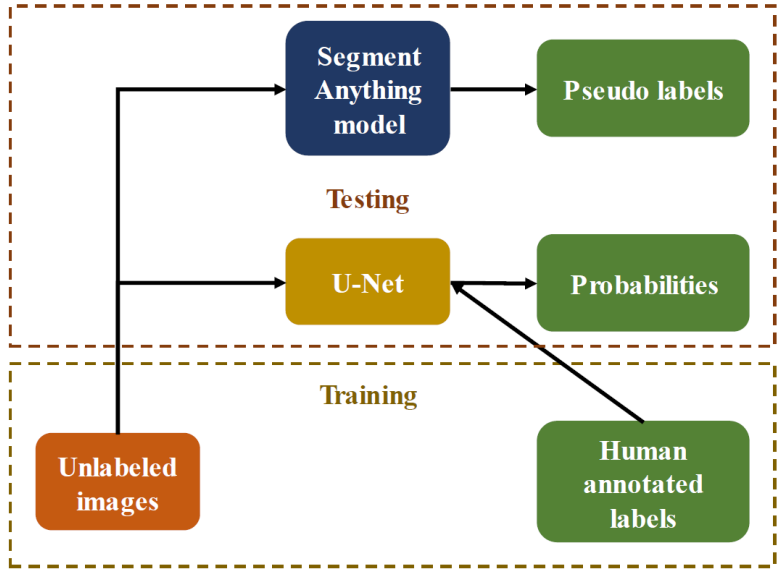

**Figure 3**. Overview of our proposed method for training cracking image segmentation and comparison the test result with the Segment Anything Model (SAM) to generate pseudo labels

Images are convolved to generate similarity scores between each area of the image and its distinctive features. After convolution, the negative similarity values of the image matrix are removed in the activation layer using the rectified linear unit (ReLU) transformation. Following the activation layer, the image matrix is condensed into a single vector in the pooling layer. This vector is fed into a fully connected neural network for classification. A correspondence score is determined for each classification label by comparing the image vectors of the training image and the input image. Classification will be indicated by the highest number. As our base segmentation frameworks, we use U-Net and SAM architectures to evaluate the performance of crack detection. Figure 3 illustrates our presented method with tasks for civil infrastructure image segmentation and comparison with SAM.

The U-Net convolutional neural network architecture was initially proposed for the segmentation of biological images. Since then, it has been modified and applied to various computer vision applications, including the identification of fractures in concrete structures. U-Net architecture is built on an encoder-decoder framework and consists of two primary components: encoders and decoders. In the U-Net design, the encoder component reduces the spatial resolution of the input image while increasing its channel count. This is achieved using convolutional layers and max pooling procedures. The max pooling procedures reduce the spatial resolution of the feature maps, while the convolutional layers extract features from the input image. The encoder component captures the context of the input image by gradually expanding the receptive field of the convolutional filters. U-Net's decoder component up-samples the feature maps produced by the encoder. To achieve this, several deconvolutional layers are concatenated with the corresponding feature maps from the encoder. The feature maps from the encoder and decoder are up-sampled by concatenating them. The decoder part of the U-Net architecture recovers the spatial resolution of the input image by combining high-level features from the encoder with low-level features from the decoder. Skip connections are used in the U-Net architecture to connect the encoders and decoders. These skip connections allow the U-Net architecture to capture both local and global features of the input image, which makes it particularly effective for image segmentation tasks. The skip connections enable the network to transfer information from the encoder to the decoder without losing the low-level features that are important for precise segmentation.

### 3.7. ResNet-18 and ResNet-50

ResNet-18 and ResNet-50 are convolutional neural network architectures introduced in 2015 as part of the ResNet (Residual Network) family of models [51]. These models use a residual block architecture and add skip connections (shortcuts) that bypass one or more convolutional layers to enable gradient flow during training, which helps in addressing the vanishing gradient problem in deep networks. ResNet-18 has18 layers containing residual blocks with two or three layers of convolution. It is relatively lightweight and can be trained faster than deeper models. ResNet-18 can be used for a variety of purposes, including object detection and image classification.ResNet-50 is more powerful and capable of handling more difficult tasks compared to ResNet-18, as it has 50 layers and a more complex residual block. It is used to identify objects, categorize images, and perform other computer vision tasks. Due to its deeper architecture, ResNet-50 can capture more sophisticated features and provide better performance on challenging problems. However, this comes at the cost of increased computational complexity and longer training times compared to ResNet-18.

### 3.8. MobileNetV2

As part of its research on embedded and mobile vision applications with constrained processing resources, Google Research developed a CNN architecture called MobileNetV2 [52]. MobileNetV2 integrates depth-separable convolutions with linear bottleneck layers to reduce parameters and computational costs while maintaining excellent accuracy. The effectiveness of this architecture has been demonstrated in several image categorization tasks as well as object recognition tasks. As a result of the separation between depthwise and pointwise convolution layers, a depthwise separable convolution block is formed, which is the main idea behind the MobileNetV2 design. Convolutions along the depth dimension are computed by the pointwise convolution, but convolutions along the depth dimension in the MobileNetV2 architecture are computed by applying a single convolutional filter to each channel. The depthwise convolution in MobileNetV2 is enhanced by the addition of 11 expansion layers and 11 projection layers. As a result of this addition, the bottleneck residual block is created, allowing for the use of low-dimensional tensors and reducing the processing effort needed. There are 17 constituent parts that make up MobileNetV2, which are followed by 11 convolutions and a global operation.

### 3.9. Xception

Xception is a pre-trained CNN architecture, developed using the Python programming language and Keras deep learning library, which is trained on the ImageNet database [53]. Xception succeeds the Inception V3 model by utilizing depthwise separable convolution over the original convolution operation used in Inception V3. Thanks to the separation of the learning space-related tasks from the learning channels-related tasks in Xception, the model size and, consequently, the computational cost has been significantly reduced [54]. The term "depth" refers to the number of convolutional or fully connected layers that extend from the input layer to the output layer.

### 3.10. Segment Anything Model (SAM)

The task in the SAM is defined as a segmentation task that leads to a pre-trained natural algorithm. The model must support flexible prompts and be ambiguity aware. These constraints are satisfied by a single design. The proposed method consists of an MAE pre-trained Vision Transformer (ViT) as the image encoder that computes an image embedding and a prompt encoder that embeds two sets of prompts: sparse and dense. Then, these two information sources are combined in a modified transformer decoder block as a lightweight mask decoder that predicts segmentation masks. The main goal of this task is to get a valid segmentation mask for at least one of the existing objects in an image when any segmentation prompts, including ambiguous prompts that could refer to multiple objects in the image. This method has shown promising results in detecting different types of cracks, including longitudinal and spalling cracks.

By leveraging the strengths of this method and other advanced technologies, the accuracy and efficiency of crack detection in concrete structures can be improved, ultimately ensuring their safety and longevity. Since segmentation masks are not readily available over the internet, a data engine is required to collect a massive dataset like SA-1B used to train the model. This data engine has been built in three major stages: a model-assisted stage, a semi-automatic stage, and a fully automatic stage. In the model-assisted manual stage, a professional annotator team labeled masks by selecting the objects in a set of images in order of prominence. This stage was performed using an interactive segmentation tool powered by the SAM. At the beginning of this stage, the SAM had been trained with common segmentation datasets, and after the availability of sufficient annotated masks, the SAM has trained again with a newly annotated dataset. In the semi-automatic stage, the main goal was to increase the diversity of masks. Confident masks were annotated by the SAM. Then, the annotators were asked to annotate less prominent objects. The model was retrained periodically to increase the accuracy and efficiency of the process. Finally, in the automatic stage, the mask annotation was fully automatic and merely done by the SAM.

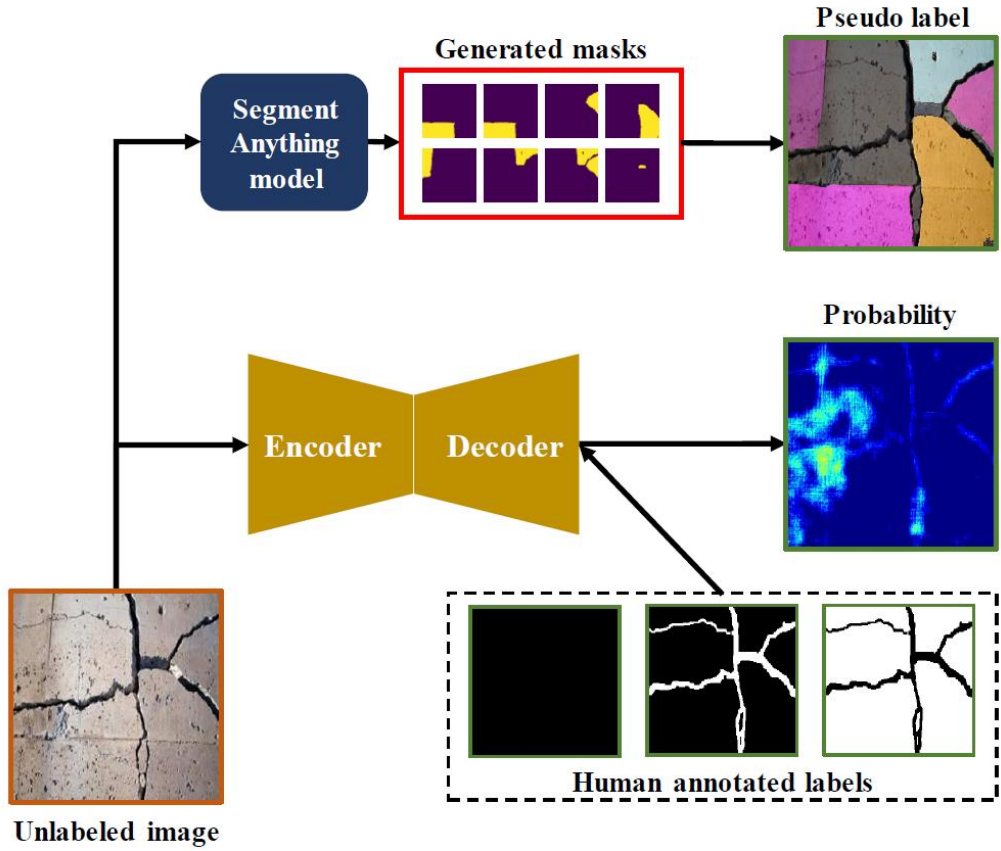

**Figure 4:** The process of crack detection using two deep learning models, SAM and U-Net.

This stage transformed the initial dataset with a limited number of masks into the final massive dataset with 11 billion masks. Throughout this stage, the model gets stronger step by step as it gets retrained periodically. Figure 4 shows that the U-Net model takes the original image as input and outputs a probability map, where the probability values represent the likelihood of a pixel being a crack. The SAM, which is pre-trained for crack detection, is used to generate labeled images indicating the location of cracks. In this study, we evaluated the performance of the SAM for crack detection in concrete structures. We used the SAM to generate segmentation results for labeled images of concrete structures with cracks. SAM generated segmented masks for each crack in the input image, which provided high prediction quality above a certain threshold. These masks were used to generate ground truth labels for our dataset. The results of this study show that the SAM is effective in detecting different types of cracks and can provide accurate segmentation masks for crack detection in concrete structures.

### 3.11. U-Net

The U-Net architecture, a CNN-based design for image segmentation, was proposed by Ronneberger et al. [55] in 2015 and has since been widely adopted for a variety of applications.. In a U-Net design, encoders and decoders play a significant role. The input image passes through a series of convolutional and pooling layers, which gradually reduce spatial resolution while increasing the number of feature maps. The encoder captures the context of the entire image and extracts high-level features from the input. The decoder, on the other hand, progressively up-samples the encoder's output, reducing the number of feature maps while restoring the original size of the encoder's output. High-level features received from the encoder are combined with precise spatial data obtained from low-level features to produce a segmentation map. During upsampling, nearest neighbors are often interpolated or convolutions are altered.

Another component of the U-Net architecture is the skip connections between the encoder and decoder at corresponding levels, allowing the decoder direct access to low-level features from the encoder. This enhances segmentation accuracy and preserves spatial information. Compared to traditional CNN systems, the U-Net architecture offers several advantages. Since both the encoder and decoder are entirely convolutional, they can handle images of any size. Furthermore, the skip connections between the encoder and decoder ensure that spatial information is preserved during the encoding process. The U-Net architecture can be efficiently trained using data augmentation techniques, even with a limited number of training samples. The U-Net architecture has been employed in numerous civil engineering applications, such as detecting building deterioration [57], identifying pavement cracks [56], and monitoring water quality. To generate multi-level features within layers, multiband inputs are filtered using the convolutional operator. Each element of the layer's feature map is computed using Equation (1).

$$\phi_{p,q,k}^l = \sum_c \sum_i \sum_j \phi_{p+i,q+j,c}^{l-1} \times w_{i,j,c,k} + b_k \tag{1}$$

To capture high-level features comprehensively, Max-pooling layers are used in addition to convolutional layers. Equation (2) concatenates the $f^{\text{UNet}}(\cdot)$ feature maps with the same size.

$$\phi^l = [\phi^{N-l+1}, \phi^{l-1}] \tag{2}$$

It should be noted that minorities within categories are weighted based on the inverse of the total counts of the categories and the total counts of pixels, i.e., minorities within categories can be given a greater weighting. In Equation (3), the loss function is calculated.

$$\mathcal{L}^{\text{seg}}(y, \hat{y}) = -(\alpha \odot y)^{\text{T}} \log \text{softmax}(\hat{y}) \tag{3}$$

The softmax function is shown in Equation (4).

$$\text{softmax}(\hat{y}) = \frac{\exp(\hat{y})}{1^{\text{T}} \exp(\hat{y})} \tag{4}$$

The input is used to make a prediction, a one-hot vector of the label is defined, and an adaptive weight vector is utilized to weight each category. Dynamically determined components make up Equation (5).

$$\alpha_i = \frac{1}{M_i} \max(M) \tag{5}$$

The adaptive weight of a category is denoted by $\alpha_i$, and the total counts of the category are shown by $i$. Gradient descent is used in the optimization process to optimize the learnable parameters in U-Net, in particular, kernels of convolutional layers. It is particularly important to note that Equation (6) calculates the derivatives of the loss function with respect to the output.

$$\frac{\partial \mathcal{L}_{\text{seg}}}{\partial \hat{y}} = \text{softmax}\,(\hat{y}) - (\alpha \odot y) \tag{6}$$

During the inference phase, U-Net generates a segmentation proposal with a size of $p \times q \times k$, suggesting that the pixels in $p \times q$ have the potential to be classified into k categories. A maximum of these k possibilities is determined by the elementwise categorization.

$$c_{p,q} = \arg \max_{k} \hat{y}_{p,q,k} \tag{7}$$

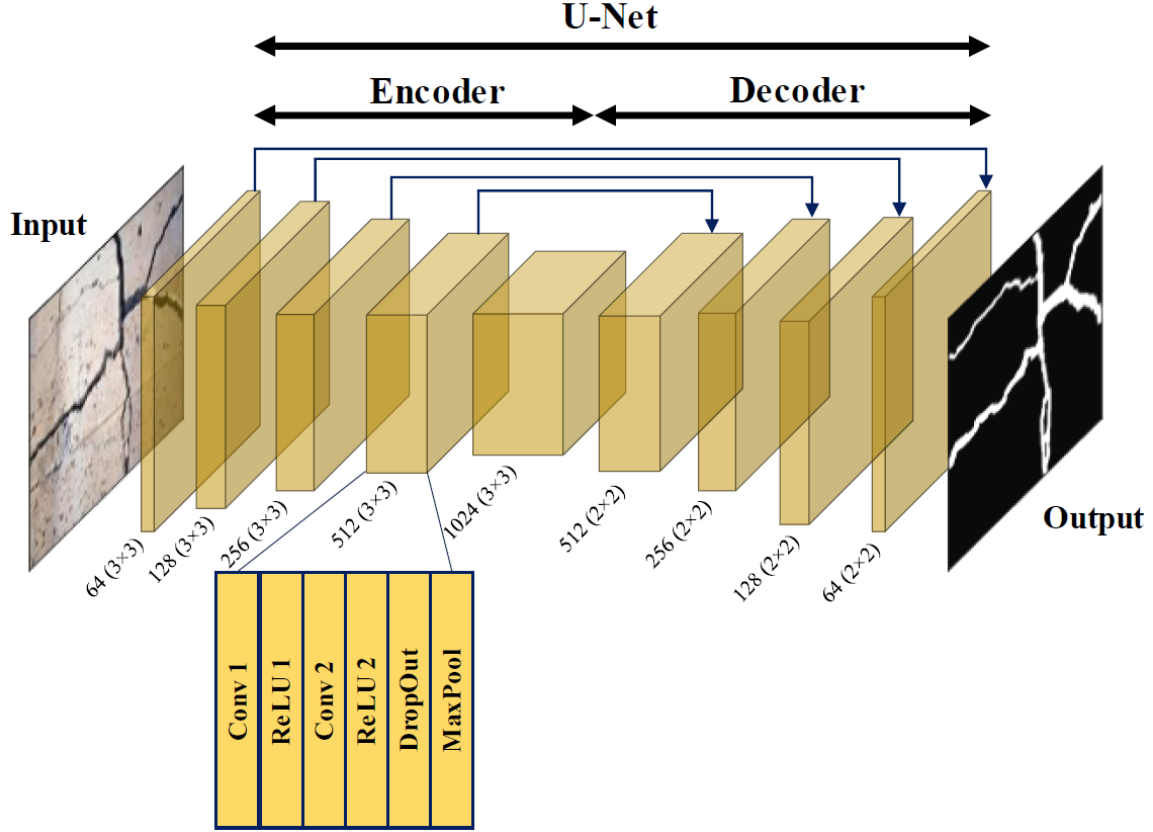

**Figure 5**: Our presented U-Net architecture

Figure 5 depicts the unique and advanced design of our proposed U-Net architecture. Table 1 also presents the composition of the layers that constitute the U-Net. This study introduces a U-Net-based deep learning model for semantic segmentation. Once the encoder has extracted high-level semantic information from the input image, the decoder generates the segmentation map and refines the spatial data. The encoder part of the U-Net comprises convolutional layers and ReLU activation layers. Each convolutional layer uses a stride of [1, 1] and a filter size of 33. This convolutional layer aims to learn the spatial representations of the input image by employing the same padding for both input and output tensors. The inclusion of ReLU activation layers in the model enables it to learn more complex functions due to their non-linearity.

**Table 2**: The layer of the U-Net structure

| Block | Stage | Layer | Type | Filter | Stride | Padding/Cropping |
|---|---|---|---|---|---|---|
| Image Input | | | 256×256×3 images with "zerocenter" normalization | | | |
| Encoder | 1 | Conv-1 | Convolution | 64 3×3×3 | [1 1] | [1 1] |
| | | ReLU-1 | ReLU | | | |
| | | Conv-2 | Convolution | 64 3×3×64 | | |
| | | ReLU-2 | ReLU | | | |
| | | MaxPool | Max Pooling | 2×2 | [2 2] | [0 0 0 0] |
| | 2 | Conv-1 | Convolution | 128 3×3×64 | [1 1] | [1 1] |
| | | ReLU-1 | ReLU | | | |
| | | Conv-2 | Convolution | 128 3×3×128 | [1 1] | [1 1] |
| | | ReLU-2 | ReLU | | | |
| | | MaxPool | Max Pooling | 2×2 | [2 2] | [0 0 0 0] |
| | 3 | Conv-1 | Convolution | 256 3×3×128 | [1 1] | [1 1] |
| | | ReLU-1 | ReLU | | | |
| | | Conv-2 | Convolution | 256 3×3×256 | [1 1] | [1 1] |
| | | ReLU-2 | ReLU | | | |
| | | MaxPool | Max Pooling | 2×2 | [2 2] | [0 0 0 0] |
| | 4 | Conv-1 | Convolution | 512 3×3×256 | [1 1] | [1 1] |
| | | ReLU-1 | ReLU | | | |
| | | Conv-2 | Convolution | 512 3×3×512 | [1 1] | [1 1] |
| | | ReLU-2 | ReLU | | | |
| | | DropOut | 50% dropout | | | |
| | | MaxPool | Max Pooling | 2×2 | [2 2] | [0 0 0 0] |
| Bridge | | Conv-1 | Convolution | 1024 3×3×512 | [1 1] | [1 1] |
| | | ReLU-1 | ReLU | | | |
| | | Conv-2 | Convolution | 1024 3×3×1024 | [1 1] | [1 1] |
| | | ReLU-2 | ReLU | | | |
| | | DropOut | 50% dropout | | | |
| Decoder | 1 | UpConv | Transposed Convolution | 512 2×2×1024 | [2 2] | [0 0 0 0] |
| | | UpReLU | ReLU | | | |
| | | | Depth concatenation of 2 inputs | | | |
| | | Conv-1 | Convolution | 512 3×3×1024 | [1 1] | [1 1] |
| | | ReLU-1 | ReLU | | | |
| | | Conv-2 | Convolution | 512 3×3×512 | [1 1] | [1 1] |
| | | ReLU-2 | ReLU | | | |
| | 2 | UpConv | Transposed Convolution | 256 2×2×512 | [2 2] | [0 0 0 0] |
| | | UpReLU | ReLU | | | |
| | | | Depth concatenation of 2 inputs | | | |
| | | Conv-1 | Convolution | 256 3×3×512 | [1 1] | [1 1] |
| | | ReLU-1 | ReLU | | | |
| | | Conv-2 | Convolution | 256 3×3×256 | [1 1] | [1 1] |
| | | ReLU-2 | ReLU | | | |
| | 3 | UpConv | Transposed Convolution | 128 2×2×256 | [2 2] | [0 0 0 0] |
| | | UpReLU | ReLU | | | |
| | | Depth concatenation of 2 inputs | | | | |
| | | Conv-1 | Convolution | 128 3×3×256 | [1 1] | [1 1] |
| | | ReLU-1 | ReLU | | | |
| | | Conv-2 | Convolution | 128 3×3×128 | [1 1] | [1 1] |
| | | ReLU-2 | ReLU | | | |
| | 4 | UpConv | Transposed Convolution | 64 2×2×128 | [2 2] | [0 0 0 0] |
| | | UpReLU | ReLU | | | |
| | | Depth concatenation of 2 inputs | | | | |
| | | Conv-1 | Convolution | 64 3×3×128 | [1 1] | [1 1] |
| | | ReLU-1 | ReLU | | | |
| | | Conv-2 | Convolution | 64 3×3×64 | [1 1] | [1 1] |
| | | ReLU-2 | ReLU | | | |
| Final Convolution | | | Convolution | 3 1×1×64 | [1 1] | [1 1] |
| Softmax | | | | | | |
| Segmentation | | Pixel Classification | | Cross-entropy loss with "C1," "C2," and 1 other classes | | |

The decoding portion of U-Net consists of transposed convolutional layers, depth concatenation layers, and ReLU activation layers. Transposed convolutional layers double the resolution of feature maps by combining semantic information from the encoder with spatial information from the decoder. As a final component, the model includes a 1x1 convolutional layer with three filters and a stride of [1,1]. 'Same' padding is used in this layer. This layer is employed to align the number of channels in the feature map with the number of classes in the segmentation. The output of this layer is then converted into a probability distribution across classes by applying a softmax activation layer.

Finally, the pixel classification layer computes the cross-entropy loss between the true labels and the predicted probability distribution. Loss-based training is used to update the network's weights. Overall, this book discussed the effectiveness of U-Net-based deep learning in semantic segmentation. Using its encoder-decoder architecture, it can refine the spatial information and produce a final segmentation map from the input image. A high-quality segmentation map is obtained by combining several convolutional layers, ReLU activation layers, transposed convolutional layers, depth concatenation layers, and softmax activation layers. In the last layer, the cross-entropy loss is calculated and used to train the model and enhance its performance

### 3.12. ADAM-Optimizer

One of the most popular techniques in the field of neural networks is the adaptive moment estimation method (ADAM), which has been used in several research studies [58, 59]. When employed with multilayer neural networks or convolutional neural networks [59], ADAM-Optimizer converges far more quickly than any other optimizer. This study segments crack images by employing Adam's optimizer for U-net's convergence application.

**Algorithm 1: ADAM-Optimizer**

Data: $\eta_t := \frac{\eta}{\sqrt{t}}$ as step size, $\beta_1, \beta_2 \in (0,1)$ as decay rates for the moment estimates, $\beta_{1,t} := \beta_1 \lambda^{t-1}$ with $\lambda \in (0,1)$, $\epsilon > 0$, $e(w(t))$ as a convex differentiable error function and $w(0)$ as the initial weight vector.

Set $m_0 = 0$ as initial $1^{st}$ moment vector

Set $v_0 = 0$ as initial $2^{nd}$ moment vector

Set $t = 0$ as initial time stamp

while $w(t)$ nol converged do

$t = t + 1$  $g_t = \nabla_w e(w(t-1))$  $m_t = \beta_{1,t} m_{t-1} + (1 - \beta_{1,t}) g_t$  $v_t = \beta_2 v_{t-1} + (1 - \beta_2) g_t^2$  $\hat{m}_t = \frac{m_t}{(1-\beta_1^t)}$  $\hat{v}_t = \frac{v_t}{(1-\beta_2^t)}$  $w(t) = w(t-1) - \eta_t \frac{\hat{m}_t}{(\sqrt{\hat{v}_t}+\epsilon)}$ end return $w(t)$

### *3.13. Loss functions*

Loss or cost functions are used to train CNN models. In order to update or optimise the weights in previous layers, the loss function, which calculates the error of segmentation or prediction, may be used. In this part, we give a brief overview of the most popular loss functions in the scientific literature. The following equations demonstrate how the prediction (or segmentation) relates to the ground truth image (or expert annotation); they index each pixel value in the image's spatial space, and the class label is represented as in classes. Cross-entropy (CE) loss is a widely used pixel-by-pixel metric for evaluating classification or segmentation performance. For problems involving two classes, CE loss functions can be expressed as Binary-CE (BCE) loss functions, as shown in the equation (8):

$$\text{Loss}_{BCE}(T, P) \qquad (8)$$
$$= -\frac{1}{N} \sum_{n=1}^{N} [T_n \cdot \log(P_n) + (1 - T_n) \cdot \log(1 - P_n)]$$

The CE loss function based on the softmax layer's output is applied to each pixel. Ronneberger et al. [55] suggested that a CE loss function with a weighting scheme could be used to improve the precision of U-Net's segmentation of cell borders in biomedical images, thereby enhancing the model's performance. A number of research on CE-based loss functions have also been done, but few of these account for the geometrical specifics of the objects [60]. Dice coefficient-based loss function (DC) that used for evaluation of the segmentation purpose is now employed due to the acceptable results. The DC calculates how much the segmentation and reference overlap. A DC of 1 denotes a total and flawless overlap (DC ranges from 0 to 1). This cost function can be defined as:

$$DC(T, P) = 2 \cdot \frac{\sum_{n=1}^{N} (T_n \times P_n)}{\sum_{n=1}^{N} (T_n + P_n)} \tag{9}$$

As a result of its tendency to provide the best segmentation, DC loss is defined in Equion (10).

$$\text{Loss}_{DC} (T, P) = 1 - DC(T, P) \tag{10}$$

Despite being good at segmenting images, CE and DC loss functions still have two significant drawbacks.

### *3.14. Performance Metric*

Traditional performance metrics, including overall classification accuracy, F-measures, and their complimentary measure, misclassification rates, are insufficient when dealing with the classification of unbalanced data. In the domain of software defect prediction, the proportion of fault-prone modules compared to non-fault-prone modules is significantly smaller. The rea under curve (AUC) is a performance metric that assesses a classifier's ability to distinguish between two classes. AUC has been demonstrated to be more dependable and to have a smaller variation than other performance measurements that were mentioned above. The field of software engineering has utilized AUC extensively.

A two-group classification problem, such as fault-prone and not fault-prone, can result in four possible outcomes: true positives (TP), false positives (FP), true negatives (TN), and false negatives (FN), where positive and negative refer to fault-prone and not fault-prone, respectively. The AUC metric ranges from 0 to 1. The ROC curve is used to determine the trade-off between the hit rate (true positive) and false alarm rate (false positive). True positive rate and False positive rate are represented by $\frac{|TP|}{|TP| + |FN|}$ and $\frac{|FP|}{|FP| + |T|}$, respectively. It is often desired that a classifier's area under the curve be used instead of one with a smaller area. Traditional performance measures, by default, only take 0.5 into account. A ROC curve displays the effectiveness at every decision threshold. The AUC is set to 1 in a perfect classifier.

## **4- Results and Discussion**
### *4-1-Results of segmentation using U-Net*

In this study, we analyzed the effectiveness of two machine learning approaches for identifying concrete cracks through image processing. First, we used the U-Net network, a well-known design for semantic segmentation tasks. The U-Net model was trained using the original images as input and the mask images as the source of truth. This network comprises an encoder-decoder architecture that is entirely convolutional, with several convolutional layers used in the encoder section to extract features from the input image. The feature maps are upsampled to the same size as the input image in the decoder section, using a succession of deconvolutional layers. During the training process, we employed a stopping criterion to prevent overfitting of the U-Net model, which occurs when the model becomes too complex and starts to fit noise in the training data instead of learning the underlying patterns. We monitored the accuracy of the model during training to prevent overfitting. If the accuracy did not improve after 10 epochs, the training process was terminated. An epoch refers to a complete pass through the data during training. By employing this stopping criterion, we ensured that the model did not overfit the training data. Figure 6 illustrates the training process.

In Figure 6, U-Net is compared with four other deep learning models, ResNet-18, MobileNet-v2, Xception, and ResNet-50, using the same dataset. For training the images, Savino et al. [26] used pre-trained deep learning models. During the training process, U-Net assigned weights to all the images using its own algorithm. According to the results, U-Net had a higher training accuracy than the other models. Therefore, U-Net's model to training images without relying on pre-trained models may result in more accurate predictions.

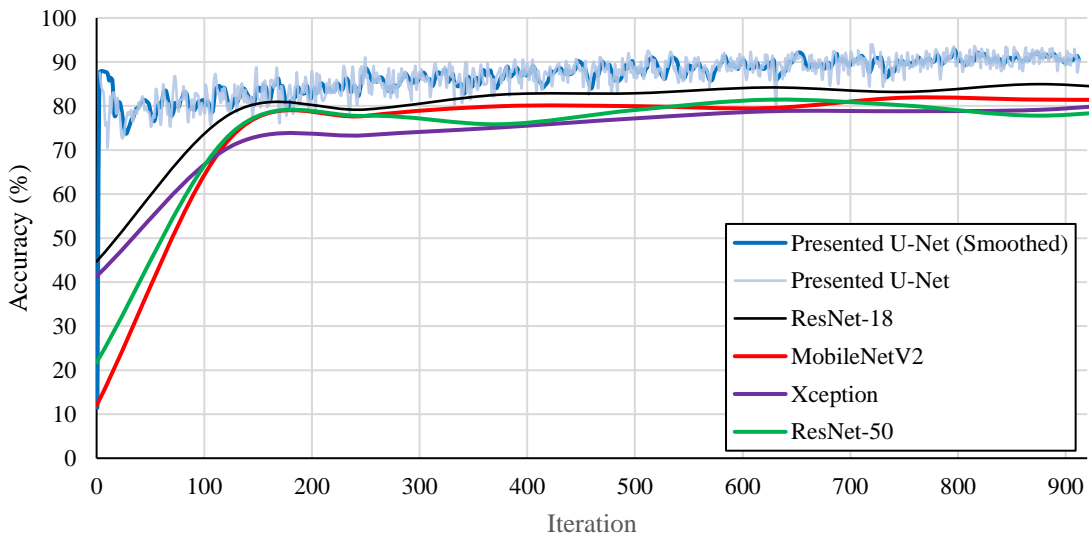

**Figure 6:** Training process of U-Net The accuracy value during the training

The U-Net model was trained for 30 epochs using RGB images with a resolution of 256x256. The backpropagation method was utilized to update the model at each epoch to minimize the loss function. The Adam method was employed for the model's optimization and convergence, with a learning rate of 0.0001 and a minibatch size of 32. Cross-validation was performed every ten epochs. Training was stopped when the accuracy showed no significant improvement in the last ten epochs. The training process resulted in an accuracy of 90% with a loss value of 0.2. Our findings demonstrate that the U-Net model can effectively detect the location and shape of specific types of cracks in concrete images. However, it is important to note that U-Net is limited to detecting cracks in concrete images and does not address the underlying causes of cracks. Identifying the root causes of cracks in concrete structures is an important area of research that can help prevent damage to structures and ensure their longevity.

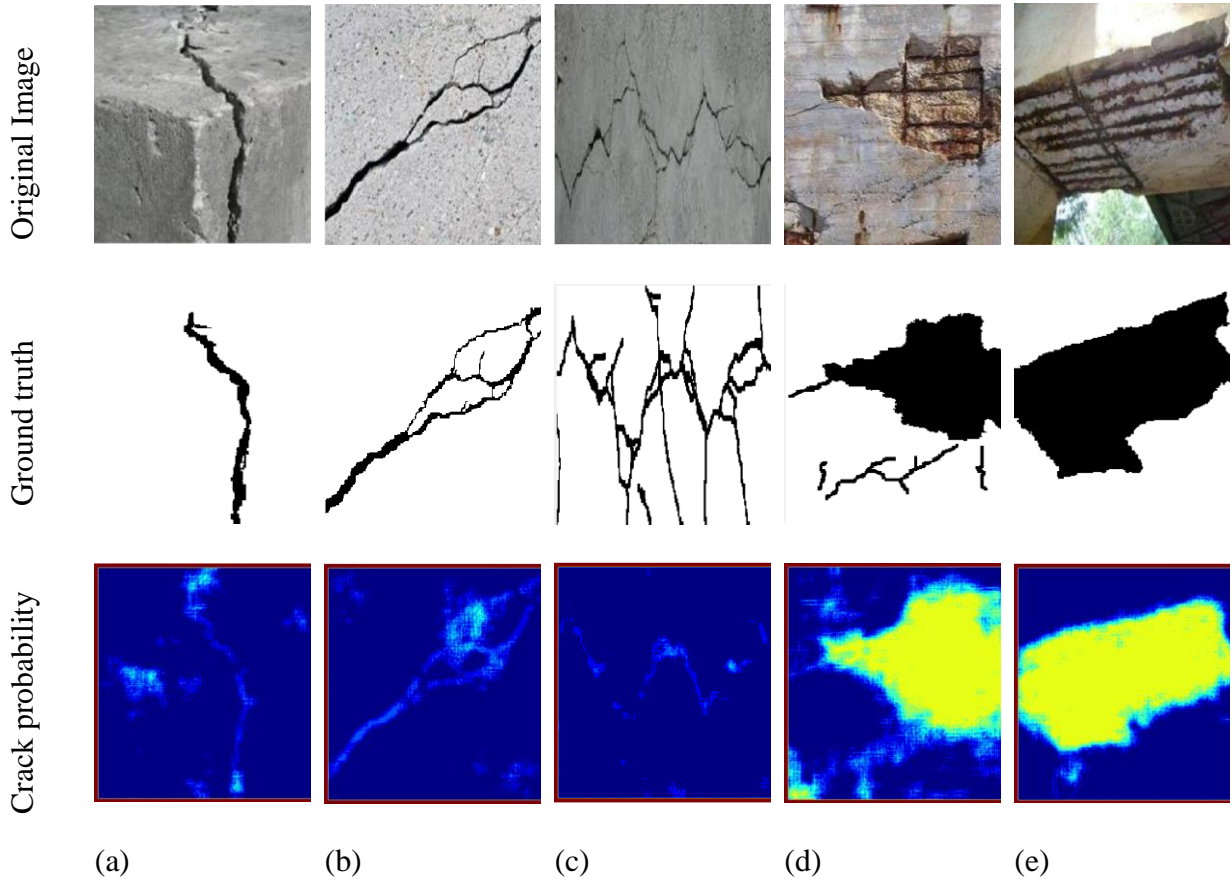

(a)          (b)          (c)          (d)          (e)

**Figure 7:** Segmentation results of U-Net a,b,c show the longitudinal crack and d and e depict the spalling cracks.

The results of the U-Net segmentation are presented in Figure 7, which shows the output of the model after the supervised learning process. The U-Net model used the dice pixel classification method, which means that classification was done for all pixels of the image. As a result, the output of the model included a spread of positive labels indicating the location of cracks in the image. To provide a more meaningful and comparable measure of the crack location, we defined a probability function based on the adjacent pixels. The purpose of this function was to determine the most probable location of cracks in the image based on the values of neighboring pixels. The probability function was defined as follows:

$$P(i,j) = 1 - \frac{\sum_{k=i-n}^{k=i+n}\sum_{l=j-n}^{l=j+n} C(k,l)}{\max C} \tag{11}$$

where P(i,j) is the probability of a crack being located at pixel (i,j), n is the number of adjacent pixels considered, and C(x,y) is the label as binary value of pixel (i,j).

To calculate the probability of a crack being located at a particular pixel, we summed the label values of the adjacent n pixels and divided them by the maximum value of the summation for all pixels in the image. We subtracted this value from one because the ground truth images of the dataset included three classes that are labeled as C1, C2, and C3. In these images, class C1 showed the crack location as zeros and the rest of the pixels as one in a binary and logical format. By subtracting the probability value from one, we were able to highlight the most probable location of the cracks in the image. The probability function provided a more intuitive and interpretable measure of the crack location in the image. By using this function, we were able to determine the most probable location of the cracks based on the values of neighboring pixels, which was a more meaningful measure than the spread of positive labels produced by the U-Net model.

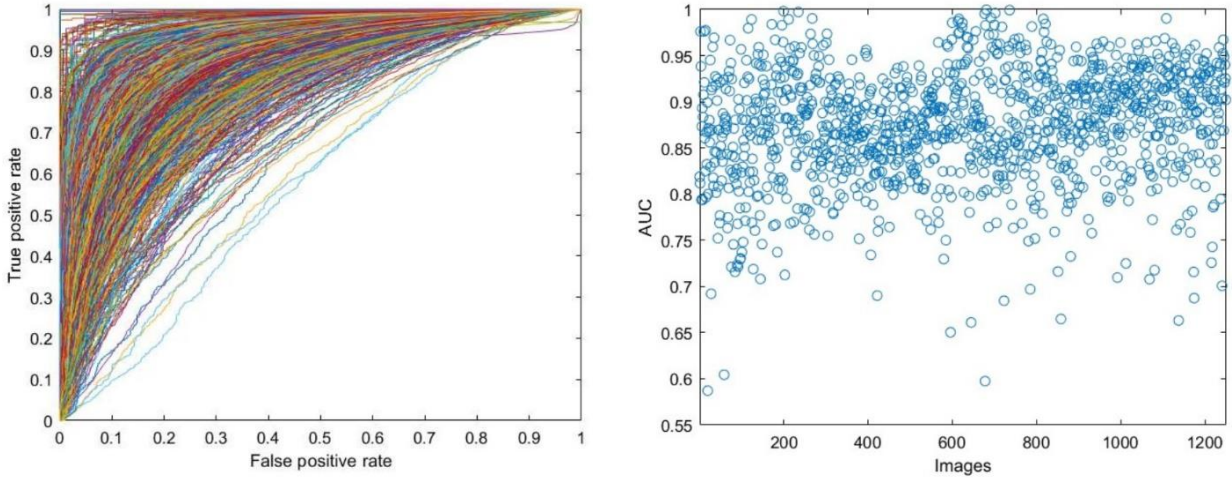

**Figure 8:** Results of segmentation using U-Net a) ROC curve, b)AUC value for each image

Figure 8a illustrates an instance of U-Net segmentation, showcasing the ROC (Receiver Operating Characteristic) curve. The ROC curve is a graphical representation of the binary classifier's performance, where the true positive rate (sensitivity) is plotted against the false positive rate (1-specificity) for different threshold values. In this case, the positive label C1 indicates the presence of a fracture in the image. A high area under the ROC curve indicates greater accuracy in locating the crack. The range of AUC values is from 0.5 to 1, with a value of 1 indicating a perfect classifier, while a value of 0.5 represents a random classifier. Each curve in Figure 8a represents the ROC curve for each image that was input into the U-Net. The AUC value for each image is depicted in Figure 8b, providing a better understanding of the results. The average AUC value for most of the images is 90%, indicating that the crack location is detected with high accuracy.

### 4-2-Results of segmentation using segment anything model(SAM)

Spalling cracks in concrete occur when the top layer of concrete flakes away, leaving the surface rough and pitted. This phenomenon can be caused by various factors, including freeze-thaw cycles, exposure to de-icing salts, and poor concrete mix design [60]. Spalling can occur in any area where concrete is used, but it is most common in areas with cold climates and frequent freeze-thaw cycles. Spalling can significantly impact the performance of concrete structures as it weakens the surface and leads to further damage. Possible spalling cracks in concrete include corrosion-induced cracking, alkali-silica reaction cracking, freeze-thaw cracking, chemical attack cracking, and abrasion cracking. On the other hand, longitudinal cracks are cracks that run parallel to the direction of the reinforcing steel in concrete slabs or walls. They can be caused by various factors, including concrete shrinkage, thermal expansion and contraction, settlement or movement of the subgrade, or overloading of the concrete [61]. Longitudinal cracks can significantly impact the performance of concrete structures, as they can weaken the structure and allow moisture and other harmful substances to penetrate the concrete. Over time, this can lead to further damage and deterioration. Prevention of longitudinal cracks involves proper design and installation of reinforcing steel, use of contraction joints and other forms of jointing to control cracking, and proper curing of the concrete. Repair of longitudinal cracks may involve filling the crack with a flexible sealant or injecting epoxy into the crack to strengthen it [62]. It is important to address longitudinal cracks as soon as they are noticed to prevent further deterioration and maintain the structural integrity of the concrete. Regular inspection and maintenance can help identify and address cracking and other forms of concrete damage [62]. In this study, we employed the Segment Anything Model (SAM) model for crack detection in concrete and compared its performance with that of U-Net. The results of the SAM network are presented in Figure 9.

The SAM model detects the surface of the crack, which is a combination of surfaces that illustrate the crack location. The technique used in SAM for crack detection is different from that of U-Net, which is trained to detect the pixels that indicate the crack. This difference provides the SAM model with higher advantages. Specifically, the SAM model is more powerful in detecting longitudinal cracks than U-Net, as it separates the image into different portions that illustrate the crack location. In contrast, the U-Net model defines the crack as positive label pixels that should be trained to detect by the model. Sometimes, when the crack is narrow, the U-Net model cannot detect the crack with high probability. However, the SAM model segments longitudinal cracks with ease. On the other hand, the U-Net model is an appropriate method for the detection of Spalling cracks in the concrete due to the shape of the crack.

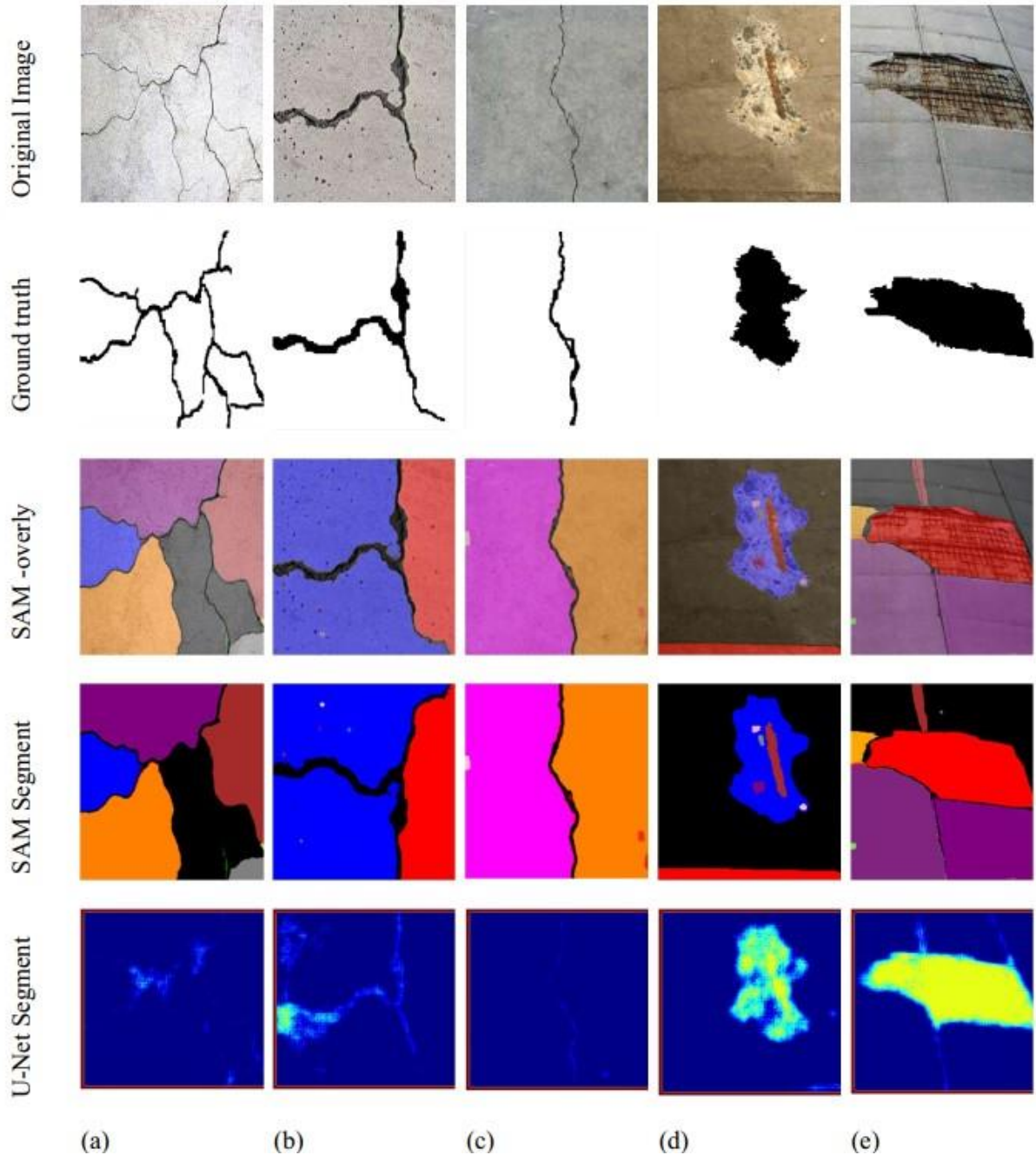

**Figure 9:** Comparison of segmentation results of SAM and U-Net

The U-Net model can detect the location and size of the spalling crack with higher accuracy. However, the SAM model does not show higher performance in detecting spalling cracks because the pre-trained SAM model was not trained to detect cracks in concrete.

## 5- Discussion

The detection of cracks in concrete structures is of utmost importance in ensuring their safety and longevity. In this study, we presented the results of using the SAM model and compared it with the U-Net model for the detection of cracks in concrete. Our results showed that the SAM model outperformed the U-Net model in the detection of longitudinal cracks while the U-Net model was better suited for the detection of spalling cracks. The U-Net model is a commonly used deep learning architecture that is popular for image segmentation tasks. It works by taking an image as input and generating an output mask that identifies the pixels that belong to a specific class or object. In the case of crack detection, the U-Net model can be trained to identify the pixels that belong to a crack in the concrete surface. This approach is particularly useful for detecting smaller cracks that may be difficult to detect with the naked eye. On the other hand, the presented SAM model is a new and innovative approach to crack detection that focuses on identifying the surfaces that make up the crack. It uses a deep learning architecture that combines a segmentation and detection network to detect the crack location. The technique used in the SAM model is different from the U-Net model because it detects the surfaces that make up the crack instead of the individual pixels. This approach can be particularly useful for detecting larger cracks and separating them into different portions for analysis.Spalling cracks are defined as the flaking of the top layer of concrete, leading to a rough and pitted surface. Our results showed that the U-Net model was better suited for the detection of spalling cracks than the SAM model. The shape of spalling cracks makes them easier to detect using the U-Net model as it can accurately detect the location and size of the crack. In contrast, the SAM model did not perform as well in detecting spalling cracks because it was not trained to detect such cracks in concrete. Our study shows that the SAM model can be an effective tool for detecting longitudinal cracks in concrete structures. These types of cracks can weaken the structure and allow moisture and other harmful substances to penetrate the concrete. Thus, early detection of longitudinal cracks is crucial in maintaining the structural integrity of the concrete. Longitudinal cracks can be caused by a variety of factors, including concrete shrinkage, thermal expansion and contraction, settlement or movement of the subgrade, or overloading of the concrete. The U-Net model, on the other hand, is better suited for the detection of spalling cracks. These types of cracks can be caused by a variety of factors, including freeze-thaw cycles, exposure to de-icing salts, and poor concrete mix design. Spalling can have a significant impact on the performance of concrete structures as it weakens the surface and leads to further damage.

The U-Net model can accurately detect the location and size of the spalling crack, allowing for early detection and repair. Both models have their advantages and disadvantages, and their suitability for a particular application depends on the nature of the cracks and the specific requirements of the project. Therefore, it is important to evaluate and compare the performance of these models in different scenarios to determine which one is more suitable for the given application. The study provides valuable insights into the performance of both models in detecting different types of cracks in concrete, which can help engineers and researchers make informed decisions about which model to use for a particular application.

## 6. Conclusion

In this study, we presented the results of using the SAM and U-Net model for crack detection in concrete. The results showed that both models have their own advantages and limitations in detecting different types of cracks. The SAM was more effective in detecting longitudinal cracks, while the U-Net model performed better in detecting spalling cracks. The SAM uses a different technique than the U-Net model in crack detection. Rather than identifying individual pixels, the SAM detects surfaces that illustrate the crack location. This method allowed the SAM to easily detect longitudinal cracks, as it separates the image into different portions that indicate the crack location. In contrast, the U-Net model was better at detecting spalled cracks due to the shape of the fracture. To determine the location and size of the spalled crack, the U-Net model identifies positive label pixels that indicate its presence. This study's findings demonstrate that the combination of SAM and U-Net model provides a more comprehensive and reliable method of identifying concrete structure fractures. SAM can detect longitudinal cracks, while the U-Net model is more effective at detecting spalled cracks.

To ensure the safety and durability of concrete structures, cracks must be located and repaired promptly. If left unattended, cracks can weaken the strength and durability of concrete and result in further damage. To maintain the integrity and safety of concrete constructions, advanced technologies such as the SAM and U-Net model described here should be employed for crack detection. The findings of this study can also have important implications for the field of civil engineering. The SAM and U-Net model can be applied to a wide range of concrete structures, including bridges, buildings, and roads. The use of these models can improve the accuracy and efficiency of crack detection, which can ultimately save time and resources in the maintenance and repair of concrete structures. In conclusion, the SAM and U-Net model presented in this study offer promising solutions for crack detection in concrete structures. While each model has its own strengths and limitations, a combination of both models can provide more accurate and comprehensive results. The findings of this study can contribute to the development of more advanced technologies for crack detection in concrete structures, which can ultimately improve the safety and longevity of these structures. The presented SAM and U-Net model for crack detection in concrete have shown promising results in this study. However, there is still room for improvement and further research in this area. One potential avenue for future work is to explore the use of different deep learning models for crack detection in concrete. There are many different types of deep learning models, and it may be possible to find a model that is better suited for this particular application. Finally, it would be useful to investigate the performance of these models on larger and more diverse datasets. The dataset used in this study contained a relatively small number of images and only a limited number of crack types. By testing the models on larger and more diverse datasets, it may be possible to gain a better understanding of their performance in different contexts and under different conditions.

**Funding Statement**

In this paper, the authors did not receive funding from any institution or company and declared that they do not have any conflict of interest.

**Code and Data available**

We would like to inform you that the code and data associated with our paper have been made available for readers on GitHub, accessible via the following link:

**https://github.com/AbbasSharifi16/SegmentAnything_CrackDetection**

We would also like to draw your attention to the preprint version of our paper, available at the following URL: **https://arxiv.org/abs/2304.12600.**